\documentclass{article}

\usepackage[nonatbib,preprint]{neurips_2021}

\usepackage[utf8]{inputenc} % allow utf-8 input
\usepackage[T1]{fontenc}    % use 8-bit T1 fonts
\usepackage{hyperref}       % hyperlinks
\usepackage{url}            % simple URL typesetting
\usepackage{booktabs}       % professional-quality tables
\usepackage{amsfonts}       % blackboard math symbols
\usepackage{amsmath}
\usepackage{nicefrac}       % compact symbols for 1/2, etc.
\usepackage{microtype}      % microtypography
\usepackage{xcolor}         % colors

\usepackage{caption}
\usepackage{subcaption}
\usepackage{graphicx}
\usepackage{soul}

\newcommand{\minjia}[1]{\sethlcolor{green} \hl{Minjia: #1}}

\newcommand{\ours}{NxMTransformer}
\newcommand{\oursspace}{NxMTransformer }
\title{NxMTransformer: Semi-Structured Sparsification for Natural Language Understanding via ADMM}

\author{%
  Connor Holmes\\
  Colorado School of Mines\\
  Golden, CO 80401 \\
  \texttt{cholmes@mines.edu}
  \And
  Minjia Zhang\\
  Microsoft\\
  Bellevue, WA 98004\\
  \texttt{minjiaz@microsoft.com}
  \AND
  Yuxiong He\\
  Microsoft\\
  Bellevue WA, 98004\\
  \texttt{yuxhe@microsoft.com}
  \And
  Bo Wu\\
  Colorado School of Mines\\
  Golden, CO 80401\\
  \texttt{bwu@mines.edu}
}

\begin{document}

\maketitle

\begin{abstract}

Natural Language Processing (NLP) has recently achieved great success by using huge pre-trained Transformer networks. However, these models often contain hundreds of millions or even billions of parameters, bringing challenges to online deployment due to latency constraints. Recently, hardware manufacturers have introduced dedicated hardware for NxM sparsity to provide the flexibility of unstructured pruning with the runtime efficiency of structured approaches. NxM sparsity permits arbitrarily selecting M parameters to retain from a contiguous group of N in the dense representation. However, due to the extremely high complexity of pre-trained models, the standard sparse fine-tuning techniques often fail to generalize well on downstream tasks, which have limited data resources. To address such an issue in a principled manner, we introduce a new learning framework, called NxMTransformer, to induce NxM semi-structured sparsity on pretrained language models for natural language understanding to obtain better performance. In particular, we propose to formulate the NxM sparsity as a constrained optimization problem and use Alternating Direction Method of Multipliers (ADMM) to optimize the downstream tasks while taking the underlying hardware constraints into consideration. ADMM decomposes the NxM sparsification problem into two sub-problems that can be solved sequentially, generating sparsified Transformer networks that achieve high accuracy while being able to effectively execute on newly released hardware. We apply our approach to a wide range of NLP tasks, and our proposed method is able to achieve 1.7 points higher accuracy in GLUE score than current best practices. Moreover, we perform detailed analysis on our approach and shed light on how ADMM affects fine-tuning accuracy for downstream tasks. Finally, we illustrate how NxMTransformer achieves additional performance improvement with knowledge distillation based methods.

\end{abstract}

\section{Introduction} \label{sec_introduction}

Large-scale Transformer networks have achieved remarkable success for a wide variety of natural language tasks, including natural language inferencing, sentiment analysis, question answering, and others. The state-of-the-art of these NLP models employs a transfer learning paradigm which contains two stages: a semi-supervised pre-training stage that trains a masked language modeling on massive web text, followed by a fine-tuning stage where the pre-trained model is adapted to specific downstream tasks with much smaller datasets. The size of these language models has dramatically increased in recent years; even relatively small models \cite{bert,distilbert} consist of hundreds of millions of parameters while larger models \cite{gpt,turing-nlg,2020t5} stretch well into multi-billions. The large model size brings challenges for both 
deployment and training costs. While training a large-scale model often requires significant time even on large training clusters, the trained model also incurs significant challenges in deployment due to latency and capacity constraints. 
% while the ability to meet strict serving latency agreements has decreased.

These challenges motivate techniques to compress and accelerate these models, even in datacenter environments where hardware resource limitations are at their smallest. These techniques include but are not limited to model quantization~\cite{8bit-bert,q-bert,binary-bert}, low rank decomposition~\cite{ladabert}, knowledge distillation~\cite{knowledge-distill,distilbert}, and model sparsification~\cite{lottery-bert,reweighted-proximal-pruning}. These compression techniques can often be combined to maximize performance gains~\cite{nn-distiller,ladabert,deep-compression-eie}.

% Knowledge distillation consists of a family of techniques that use additional information from a large parent model, such as intermediate or logit losses, to better train a small model to achieve higher accuracy than it otherwise would. Model quantization reduces the precision of model parameters from single precision floating point values to reduced formats such as half, bfloat16, or small integer representations. 
% \minjia{Move some of the discussions of each technique to the background section.}
Among different techniques, sparsification attempts to identify parameters that can be removed from the model without significantly compromising model accuracy. Sparsification techniques typically fall under two broad categories: unstructured and structured. Unstructured techniques will remove the individual parameters based on their importance (e.g., weight magnitude), which often yield the best accuracy but are unfriendly to modern hardware. Structured sparsification techniques remove parameters in groups (e.g., entire rows or columns), which result in models that retain their dense structure but can also add constraints that limit the expressiveness of the model.
% to non-negligible accuracy drop. 

Recently, hardware manufacturers introduced support for NxM semi-structured sparsity to provide the benefits of both structured and unstructured sparsity. In NxM semi-structured sparsity, a model may preserve M parameters from each contiguous group of N original parameters. This relatively weak constraint on sparsification allows for sparse representations similar in flexibility to those of unstructured approaches but also permits efficient hardware implementation as well. Consider the 4x2 semi-structured sparsity implementation found on NVIDIA GPUs based on the Ampere architecture~\cite{nvidia-a100}. The Ampere architecture introduces a small set of multiplexers that select values from the input matrix corresponding to the retained values in the weight matrix \cite{nvidia-nxm}. The output of this operation remains compatible with the efficient Tensor Cores for dense matrix-matrix operations. While the Ampere GPUs are the first to market with this capability, the matrix multiplication accelerators within them are similar to those used by other accelerators \cite{tpu-paper} which should enable other vendors to provide support for this type of sparsity as well.

To induce semi-structured sparsity, a small set of approaches have been offered. ASP \cite{nvidia-nxm} proposes training the dense network until convergence, using single-shot magnitude-based pruning to induce sparsity conformant to the NxM constraints, and repeating the original training procedure to recover accuracy. Zhou, et al. \cite{sr-ste} uses sparse-refined straight-through estimator (SR-STE) to introduce the sparsity throughout the entire training process. Both of these techniques pose some challenges for large-scale pre-trained Transformer networks in particular. ASP can require a second costly sparse pre-train of the model and the single-shot magnitude-based pruning might hurt the knowledge transferrability to different downstream tasks. SR-STE on the other hand sparsifies from the random model initialization, which avoids the costly sparse retraining but also necessitates performing the pre-training process with only a sparse representation in mind. Since NxM sparse hardware is not yet ubiquitous, maintaining compatibility with a single dense pretrained representation is valuable so the costly pre-training process does not need to be performed for both sparse and dense networks.

\begin{figure}[!t]
    \centering
    \includegraphics[width=\textwidth]{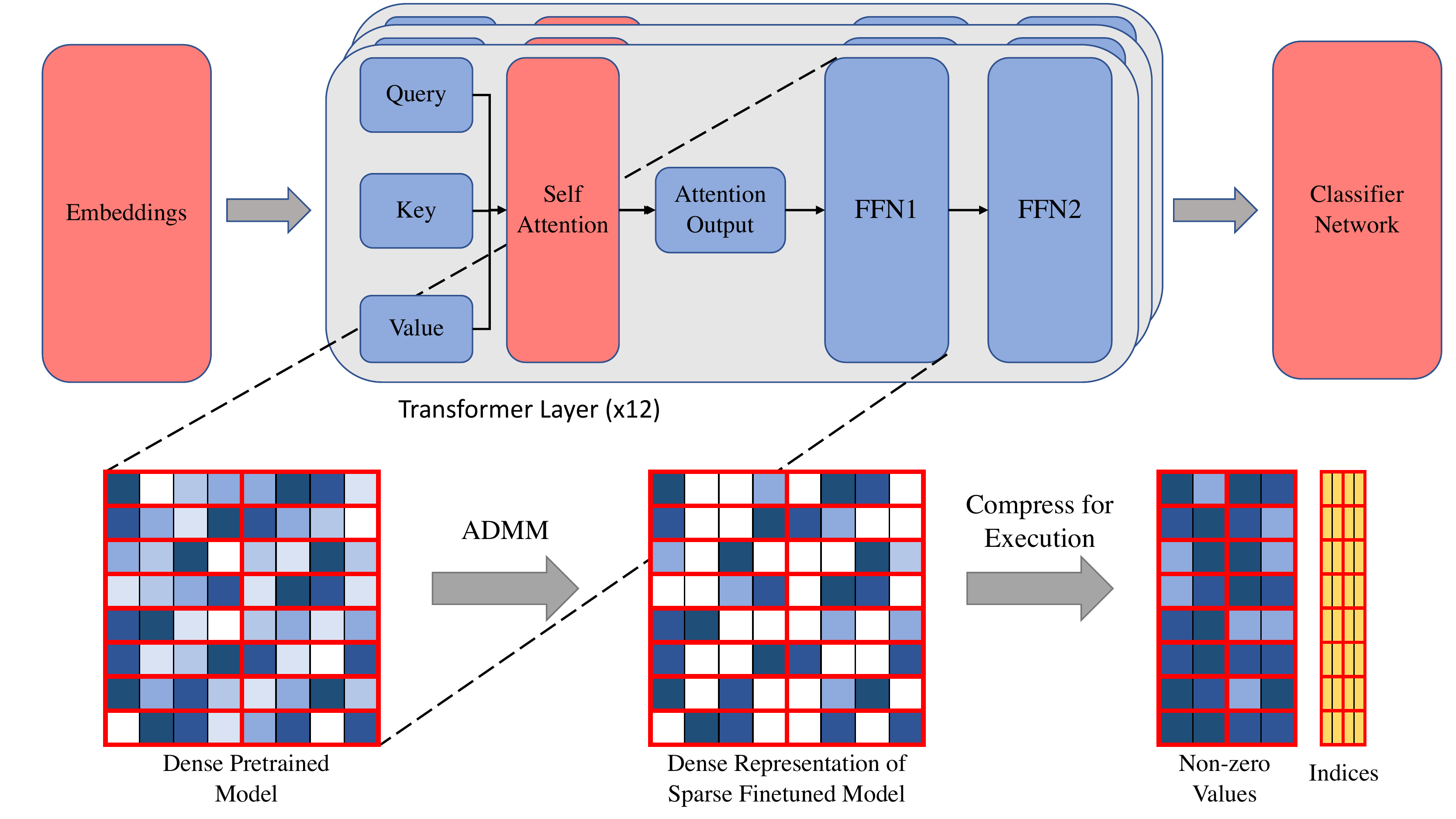}
    \caption{The layers sparsified by \oursspace are highlighted in blue in the block structure of a BERT model. As shown for FFN1, \oursspace simultaneously finetunes the pretrained representation while inducing NxM semi-structured sparsity using ADMM. This sparse model can be trivially converted to the deployment format for compatible hardware.}
    \label{fig_system-overview}
\end{figure}

\iffalse
\minjia{It is a bit vague to say "ASP does not intelligently optimize for sparsity". }
SR-STE on the other hand requires training an N:M semi-structured network from scratch, which might be feasible for small models or datasets but has its limitation of applicability to large-scale Transformer models, where pre-training is prohibitively expensive.
\fi

In this work, we propose to effectively induce NxM semi-structured sparsity for large-scale Transformer networks to leverage newly released Sparse Tensor Core hardware by making the following contributions. (1) We introduce a principled method, \oursspace (See Figure \ref{fig_system-overview}), to obtain Transformer networks with NxM semi-structured sparsity for different downstream tasks using Alternating Direction Method of Multipliers (ADMM), a technique designed for large-scale non-convex optimization problems with constraints. Such a method allows us to alternate promoting the NxM sparsity of the network and optimizing the fine-tuning performance. (2) We conduct comprehensive experiments and demonstrate that \oursspace achieves 1.7 points higher accuracy than state-of-the-art techniques to introduce NxM sparsity for natural language processing tasks. (3) We perform detailed analysis on our approach and shed light on how ADMM affects fine-tuning accuracy for downstream tasks. (4) Finally, we show that \oursspace is complimentary to alternative model compression techniques such as knowledge distillation and can further sparsify distilled models while preserving accuracy. 
% in cooperation with densely trained models compressed with knowledge distillation techniques.
% Most previous work on ADMM has been focusing on inducing both structured and unstructured sparsity into convolutional neural networks in computer vision tasks~\cite{}\text{Citation.}. Distinctive from those studies, we apply ADMM to fine-tuning pre-trained Transformer networks on NLP tasks.
% Rather than fine-tuning the model before inducing sparsity, the NxMTransformer pipeline (See Figure \ref{fig_overview}) simultaneously optimizes for sparsity while performing the fine-tuning process. Source code for this work can be found at (INSERT LINK).

% In summary, this paper makes the following three contributions. First, we demonstrate that \oursspace provides higher accuracy than competing techniques to introduce NxM sparsity for natural language processing tasks. Second, we perform the first investigation into the mechanisms by which ADMM introduces sparsity and provide guidelines for tuning ADMM's parameters for unseen downstream tasks. Finally, we demonstrate that \oursspace can sparsify in cooperation with densely trained models compressed with knowledge distillation techniques.

\section{Background and Related Work} 
\label{sec_related-work}

\paragraph{Compression Techniques for NLP Models.} 

Due to the prominence of large-scale language models, there has been significant interest in compressing these models. Sparsification has been shown to be a promising approach to improving model inference speed and reducing memory cost earlier in computer vision~\cite{deep-compression-eie}. Sparsifying pre-trained language models turns out to be a challenging task because the semi-supervised models tend to have less intrinsic sparsity than CNN-based models.
Some prior studies try to understand and apply sparsification for large-scale language model compression. 
% work has emerged into understanding the intrinsic sparsity of these networks as well as compression mechanisms. 
Chen, et al. \cite{lottery-bert} extend the Lottery Ticket Hypothesis \cite{lottery-hypothesis} to pre-trained models, finding that winning tickets for the pre-training tasks do transfer universally to downstream tasks. Gordon, et al. \cite{weight-pruning-transfer-learning} find that the level of sparsity discovered during the pretraining process represents most of the sparsity that can be discovered in the model. Guo, et al. \cite{reweighted-proximal-pruning} show that a proximal pruning strategy achieves higher accuracy than competing lasso regularization methods and iterative magnitude pruning. Most of these studies still focus on unstructured sparsity, which encounters difficulty to obtain large speedups on modern hardware. As a structured technique, Michel, et al. \cite{sixteen-heads} observe that many transformer heads themselves are redundant and can be pruned with minimal accuracy loss. While a structured technique, this technique provides limited acceleration benefits since much of the compute time for short-to-medium sequence lengths is spent in the intermediate layers rather than the attention heads themselves. 
In addition to parameter sparsification, BERT models can also realize compression gains with model quantization~\cite{8bit-bert,q-bert} and knowledge distillation~\cite{distilbert,patient-kd,mobilebert,tinybert}. These techniques are complimentary to our proposed method and can be combined to achieve more effective overall compression, as in \cite{deep-compression-eie}.

% Zafrir \cite{q8bert} successfully quantize BERT to 8-bits with minimal accuracy degradation using symmetric linear quantization. Shen \cite{qbert} use second-order Hessian information about fine-tuned models to perform mixed-precision quantization to achieve up to 13X compression of model parameters. Quantization techniques are typically orthogonal to the above sparsification techniques and can combine their benefits.

% \paragraph{Knowledge Distillation for NLP Models.} 

% Knowledge distillation uses a large fully-trained model to teach a small, compact model to learn its behavior rather than the task itself. Sanh, et al. \cite{distilbert} train the six-layer DistilBert that maintains 97\% of the accuracy of the BERT base teacher while providing 60\% inference improvement using distillation at the logits and an initialization of every other layer of the teacher model. Sun, et al. \cite{mobilebert} modify the architecture of BERT to use a narrow deep structure that's more efficient for mobile devices and combine it with logit, feature map, and attention head distillation to achieve 4.3X compression and 5.5X inference improvement on mobile hardware. Jiao, et al. \cite{tinybert} perform knowledge distillation during both the pre-training and fine-tuning phases while using projections to distill between layers of different dimensions, providing up to 7.5X compression and 9.4X inference performance improvement while maintaining 96.8\% of the teacher's accuracy.

\iffalse
\minjia{No need to talk about quantization and KD in details as they are largely orthogonal.}
\fi

\paragraph{Semi-structured Sparsity.} 

There have been few studies to induce NxM sparsity for DNN models. Among them, SR-STE \cite{sr-ste} is the most comparable work to this paper. SR-STE is single stage compression technique that uses an extended Straight-through Estimator and a sparse-refined term to improve the induction of NxM sparsity in the network. SR-STE is demonstrated across image and neural machine translation tasks and shows improved accuracy over NVIDIA's ASP \cite{nvidia-nxm}. Unlike this work, SR-STE is designed for training from random initialization, not a pretrained model.

From a performance perspective, cuSPARSElt \cite{nvidia-nxm} (the NVIDIA linear algebra library for NxM sparsity) demonstrated 1.3X to 1.6X performance uplift for the matrix multiplications prevalent in BERT models using sparse NxM Tensor cores on Nvidia A100 GPUs. Higher speedups may be achieved by increasing batch size, sequence length, or model dimensionality with a maximum of 1.8X improvement for fp16 and 1.9X for int8. Realized speedups are smaller than the theoretical 2X improvement of 50\% sparsity primarily due to memory bandwidth constraints. The performance provided by cuSPARSElt is a reasonable expectation for what can be achieved with \ours.

Models including NxM sparsity may be deployed for serving using optimized frameworks like TensorRT \cite{nvidia-tensorrt}, which include optimized layer operators that take advantage of the hardware's features. Moving forward, inference-driven machine learning compilers like TVM \cite{tvm} can also introduce support for the semi-structured sparse Tensor operators using the same mechanisms as were used to provide support for dense Tensor cores, which use a similar API.

\paragraph{ADMM for Neural Networks.} 

Alternating Direction Method of Multipliers (ADMM) has been used in previous works primarily for compressing convolutional neural networks. Ye, et al. \cite{progressive-admm} explore an iterative approach to ADMM that shortcuts full ADMM convergence with a masked retraining operation. Ren, et al. \cite{admm-nn} use ADMM for both unstructured sparsification and quantization of the model while combining the technique with a heuristic for determining whether speedup will be achieved given the achieved compression of a layer. Ma, et al. \cite{pconv} use domain knowledge of CNN filter structure to perform a structured pruning operation for mobile-hardware efficient compression. While the above techniques all use ADMM as the sparsifying mechanism, none examine in-depth applicability to pre-trained language models for NxM sparsity.

\section{Methodology} \label{sec_methodology}

In this section, we formulate ADMM for NxM sparsity, describe the specific aspects of pre-trained models that define the optimization pipeline, and describe the high-level optimization schedule.

\subsection{Problem Definition.} 
\label{subsec:problem}

Consider adapting a pre-trained large-scale language model  $\Phi$ with $L$ Transformer layers (e.g., BERT) to natural language understanding tasks, such as sentiment analysis, entailment, question and answering, etc, where the training instances are inputs (often text phrases) and target pairs: $\{x_i,y_i\}_{i=1}^N$. Assume the collection of pre-trained model weights is $W^0=\{\mathbf{W^0}_i\}_{i=1}^{L}$\footnote{For convenience's sake, we will omit the notation of the bias since it is not relevant to the task of sparsification.}, the goal of NxMTransformer is to load $W^0$ and fine-tune it to $W'$ such that  \{$W'_i\}_{i=1}^L$ satisfies the constraints of at most $M$ weight parameters having non-zero values out of $N$ consecutive weights, while achieving similar performance in comparison to fine-tuning the task-specific objective function $f(\{\mathbf{W}_i\}_{i=1}^{L})$ (e.g., cross-entropy for classification) using $W_0$ but without constraints. 

\subsection{\ours}
\label{subsec:methodology}

\iffalse
~\cite{sgd}
\fi

Different from conventional DNN training objectives, the above problem is non-convex with combinatorial constraints. Therefore, it cannot be directly solved by gradient-based methods such as stochastic gradient descent. To address this issue, we adopt the alternating direction method of multipliers (ADMM), which is a reliable method for large-scale constrained optimization (e.g., with combinatorial constraints). In particular, we modify the objective function of the NxM sparsity problem as

\begin{equation} \label{eq_admm}
\begin{aligned}
&\min_{\{\mathbf{W}_i\}} f(\{\mathbf{W}_i\}_{i=1}^{L}) + \sum_{i=1}^{L} g_i(\mathbf{Z}_i) \text{ subject to } \mathbf{W}_i = \mathbf{Z}_i, i=1,\dots,L
\end{aligned}
\end{equation}

where $f(\cdot)$ is the fine-tuning objective function, and $g_i(\cdot)$ is an added penalty function and $Z_i$ are auxiliary variables. To apply ADMM, we define the penalty function as

\begin{equation} \label{eq_indicator}
g_i(\mathbf{W}_i) = 
\begin{cases}
0 & \text{if $\mathbf{W}_i \in \mathbf{S}_i$} \\
\infty & \text{otherwise}
\end{cases}
\end{equation}

where $\mathbf{S}_i$ represents the constraint set $S_i = \{W_i$ that have at most M weight parameters having non-zero values out of N consecutive weights$\}$. 

\iffalse
\begin{figure}
    \centering
    \includegraphics[width=\textwidth]{ADMM-DS/figures/transformer-model/transformer-model.pdf}
    \caption{Basic structured of BERT model. Layers shaded in blue are sparsified by \ours.}
    \label{fig_transformer-model}
\end{figure}

\begin{figure}[t]
    \centering
    \includegraphics[width=\textwidth]{ADMM-DS/figures/nxm-overview/overview-cropped.pdf}
    \caption{System pipeline for NxMTransformer. NxMTransformer simultaneously finetunes the pretrained representation while inducing semi-structured sparsity. This format can be trivially converted to the deployment format for compatible hardware.}
    \label{fig_overview}
    \minjia{Can we combine Figure 1 and Figure 2? For example, stack these two figures, and have a zoom out view of one of these dense layers, such as FF1. The zoomed out layer is the "dense pretrained model". The rest can stay the same.}
\end{figure}
\fi

\paragraph{Choice of $S_i$.} 

Not all weights in pre-trained Transformer models need to satisfy this NxM sparsity constraint. BERT and similar pretrained models typically consist of three major components: embeddings, Transformer blocks, and classifiers (See Figure \ref{fig_system-overview}). For NxM semi-structured sparsity, we solely consider weights in Transformer blocks. Take BERT as an example, each Transformer block consists of 6 fully connected sub-layers: the query $Q$, key $K$, value $V$ layers, the attention output layer $Attn.$, and two feed-forward network $FFN1$ and $FFN2$. Each of the fully connected layers can take advantage of NxM semi-structured sparsity; furthermore, these layers constitute the vast majority of inference wall-time. Of the six fully connected layers, $FFN1$ and $FFN2$ are particularly important to sparsify, alone requiring more than half of the inference wall-time for a Transformer block. Note that attention head pruning techniques~\cite{sixteen-heads} are unable to sparsify $FFN1$ and $FFN2$. The self-attention mechanism itself does not include any trained parameters and is unable to be sparsified using this technique. We exclude the embedding layer since the lookup operation associated with the embedding layers is incompatible for acceleration with Tensor Cores. The classifier is composed of fully connected layers. 
% For sequence classification tasks, a stack of a single pre-classifier and classifying fully connected layers is attached to the leading [CLS] token. For token-level tasks, these are attached to each token of the sequence. For both of these classes of tasks, 
For a given task, the classifier weights are randomly initialized at the beginning of the fine-tuning process. We find that 
% This conflicts with the above expectation of ADMM;
sparsifying these matrices using ADMM will unnecessarily harm accuracy. Since the execution of these layers is typically under 2\% of the inference wall-time, the runtime cost is minimal for doing so.

\paragraph{Decomposing the minimization problem into sub-problems.} 

Once we define $S_i$, we apply the augmented Lagrangian, which decomposes equation~\ref{eq_admm} into two sub-problems on $W$ and $Z$:

\begin{equation} \label{eq_first-subproblem}
	\text{Sub-problem 1 (\textbf{performance-promoting}):} \min_{\{\mathbf{W}_i\}} f(\{\mathbf{W}_i\}_{i=1}^{L}) + 
	\sum_{i=1}^{L} \frac{\rho}{2} \Vert \mathbf{W}_i - \mathbf{Z}_{i}^k + \mathbf{U}_{i}^k \Vert_{F}^{2} 
\end{equation}

\begin{equation} \label{eq_second-subproblem}
	\text{Sub-problem 2 (\textbf{NxM sparsity-promoting}):} \min_{\{\mathbf{Z}_i\}} \sum_{i=1}^{L} g_i(\mathbf{Z}_i) + 
	\sum_{i=1}^{L} \frac{\rho}{2} \Vert \mathbf{W}_{i}^{k+1} - \mathbf{Z}_{i} + \mathbf{U}_{i}^k \Vert_{F}^{2} 
\end{equation}

The first sub-problem solves the \emph{performance promoting} problem, which consists of two terms. The first term is the standard objective function for fine-tuning the task, and the second term is a $L_2$ regularization term. The regularization target $Z_i^k - U_i^k$ is dynamically updated, based on $U_i^k$, which is the dual variable (i.e., the Lagrange multiplier). Since the $L_2$ term is convex, the complexity of solving sub-problem 1 (e.g., via ADAM \cite{adam}) is the same as minimizing $f(\cdot)$.
% , and can be solved effectively using a traditional gradient descent solver, such as Adam (CCITE) or any of its variants.
The second sub-problem solves the \emph{sparsity promoting} problem. Since it optimizes the sparse constraints separately, it can be solved analytically as the solution of the Euclidean projection of $\mathbf{W}_{i}^{k+1} + \mathbf{U}_{k}$ onto our constraint. For the case of NxM semi-structured sparsity, this is accomplished by retaining the M largest values of the contiguous group of N values (See Figure \ref{fig_system-overview}), which can be solved in linear time.
% $\mathbf{U}$ is an auxiliary variable updated as in the previous literature (CCITE sort of feeling like it's equation spam):
Finally, we need to update the dual variable $U$ as $\mathbf{U}_{i}^{k} := \mathbf{U}_{i}^{k-1} + \mathbf{W}_{i}^{k} - \mathbf{Z}_{i}^{k}$ to guarantee that the dual feasibility condition is satisfied in each ADMM iteration.

\paragraph{Sparsity-inducing based fine-tuning.} The typical ADMM pipeline fully trains a model to convergence before introducing ADMM \cite{admm-nn}. This two-step process is necessary since the primary objective of ADMM is to optimize the existing network to conform to the constraints; ADMM will only introduce small changes to parameters it retains in the model. However, for pre-trained language models, the primary purpose of the fine-tune is to adapt the classifiers for the specific downstream with minimal disturbance to the parameters of the pre-trained representation. As a result, we apply the three aforementioned steps (i.e., two sub-problems and the update of the dual variable) while fine-tuning the model. In particular, we perform the three steps in an alternating manner, i.e., performing some number of fine-tuning steps with Adam to solve the first sub-problem, solving the Euclidean projection for each weight matrix for the second sub-problem, and finally updating the auxiliary variable. This sequence will be referred to as one ADMM iteration. The optimization proceeds until the $\mathbf{W}$ and $\mathbf{Z}$ variables have converged, at which point we have a sparsified network compliant with our NxM constraint.

\section{Evaluation} 
\label{sec_results}

In this section, we evaluate NxMTransformer and show its effectiveness in compressing Transformer networks over a wide range of NLP tasks.

\paragraph{Implementation.} NxMTransformer is implemented as a PyTorch \cite{pytorch} compatible library for sparsifying models with NxM semi-structured sparsities.
% with user-defined ADMM constraints. 
% The released library includes built-in extensions for both unstructured and NxM semi-structured sparsities. 
Furthermore, a HuggingFace Transformers \cite{transformers-huggingface} compatible Trainer is implemented to enable easy integration with their model collection and training scripts. Our approach supports different NxM sparse patterns (e.g., 4:1, 8:4) so long as the weight's input dimension is a multiple of N. For evaluation, we focus on evaluating 4:2 sparsity since it is supported in commodity hardware.  We use pretrained model checkpoints for both BERT\footnote{\url{https://huggingface.co/bert-base-uncased}, Apache 2.0 License} and DistilBERT\footnote{\url{https://huggingface.co/distilbert-base-uncased}, Apache 2.0 License}, provided by the HuggingFace model repository. All models were fine-tuned on an Intel Xeon 2630 v4 server with 2x NVIDIA Titan V running Ubuntu 18.04. PyTorch version 1.7.1 was used alongside Transformers 4.3.2. Finetuning these models required between 5 minutes (RTE) and 5 hours (MNLI) depending on task size.
For the set of training hyperparameters used for training
% ADMM\textsubscript{NxM} with 
\ours, see Table \ref{tab_admm-hyperparameters}.

\paragraph{Dataset.} We evaluate \oursspace and our baselines using the the General Language Understanding Evaluation (GLUE) benchmark \cite{glue}, a collection of NLP tasks varying in data availability and complexity. We report the Spearman correlation for STS-B, the F1 score for MRPC, Matthews correlation for CoLA, and accuracy for all remaining tasks. The reported average is the geometric mean of reported scores.

\subsection{Main Results} \label{sec_baseline}

\iffalse
To evaluate the effectiveness of the proposed approach, we compare \oursspace with the following baselines: 
\minjia{On a second thought, I feel we should call our technique just one name, e.g., \oursspace consistently throughout the paper to avoid confusion. Could you please change the other places that use ADMM$_{NxM}$ to NxMTransformer, including text, tables, and figures? }
\fi

\begin{itemize}
    \item \textbf{BERT}~\cite{bert}: This is the BERT\textsubscript{base} model from publicly available checkpoint. 
    \item \textbf{ASP}: Inline with ASP practices\cite{nvidia-nxm}, we perform one-shot magnitude-based masked pruning on the pretrained model. This baseline is considered best practices for a large pretrained language representation for semi-structured sparsity. 
    \item \textbf{ADMM\textsubscript{Unstructured}}: To measure the accuracy cost of semi-structured accuracy specifically, we create another baseline that uses ADMM but induces unstructured sparsity at 50\%  per-layer (rather than global) sparsity.
\end{itemize}

\iffalse
\minjia{Need to separate the discussion of BERT and DistilBERT.}
\fi

\paragraph{Hyperparameters.} In \cite{bert}, the authors only report the development results on a few tasks. Therefore, we produce the BERT baseline results. We fine-tune BERT for 5 epochs on each downstream task. We perform a grid search of batch sizes 16 and 32, and learning rates 1e-5, 3e-5, and 5e-5 for SST-2, QNLI, and MNLI, due to their high training cost. Learning rates of 7e-5 and 9e-5 are additionally used for the remaining tasks.
For masked fine-tune, the model was fine-tuned with learning rates 1e-5, 3e-5, 5e-5, 7e-5, and 9e-5 across batch sizes 16 and 32. ADMM\textsubscript{Unstructured} is trained using the same hyperparameters sweeps as \ours. For all configurations, we set the fine-tune to have 5 epochs, and the best observed result on the validation set is reported.

\begin{table}[t]
\caption{
The dev set results on the GLUE benchmark. The results show that \oursspace is able to achieve higher accuracy than ASP for NxM sparsity, especially when the downstream tasks have low data resources. 
}
\label{tab_glue-results}
\resizebox{\textwidth}{!}{
\begin{tabular}{@{}lcccccccc@{}} \toprule
Model                                   & \multicolumn{7}{c}{Task} & Average \\ \cmidrule(r){2-8}
                                        & MNLI (m/mm)   & SST-2     & QNLI  & CoLA  & STS-B & MRPC  & RTE   &           \\ 
Samples                                 & 392k          & 67k       & 108k  & 8.5k  & 5.7k  & 3.5k  & 2.5k  &           \\ \midrule
Baseline (BERT\textsubscript{base})                & 84.5/84.8     & 92.5      & 91.6  & 56.7  & 89.6  & 91.7  & 70.7  & 81.8      \\
ADMM\textsubscript{Unstructured}   & 84.0/84.7     & 92.5      & 91.0  & 57.5  & 89.6  & 90.5  & 68.2  & 81.3      \\ \midrule\midrule
ASP                                & \textbf{83.3}/83.4     & 91.9      & \textbf{90.6}  & 51.7  & 88.7  & 88.1  & 63.9  & 78.8      \\
NxMTransformer            & 82.3/\textbf{83.4}     & \textbf{92.3}      & {90.4}  & \textbf{55.3}  & \textbf{89.3}  & \textbf{90.8}  & \textbf{68.6}  & \textbf{80.5}      \\ \midrule
% DistilBERT                              & 82.4/82.5     & 90.9      & 89.1  & 53.4  & 86.6  & 89.6  & 63.5  & 78.5      \\
% ADMM\textsubscript{NxM} DistilBERT      & \textbf{80.7/81.2}     & \textbf{90.5}      & \textbf{87.5}  & \textbf{50.1}  & \textbf{87.1}  & \textbf{88.7}  & \textbf{59.2}  & \textbf{76.6}      \\ \bottomrule
\end{tabular}}
\iffalse
\minjia{Our MNLIM-m accuracy is lower than ASP by 1 point?}
\textcolor{red}{That was what we ended up seeing with it. My takeaway from this (and I don't think I emphasized this enough) was basically that a bunch of training data can make up the difference between ASP and \ours.
\minjia{But our approach theoretically should not give worse results than ASP. If we set the penalty coefficient $\rho$ to be a huge number, isn't that equivalent to masked fine-tune? Also, I see we get worse results on MNLI and QNLI, which happen to also be the two benchmarks we did not tune on larger learning rates. Can we quickly try these two datasets, or at least MNLI on learning rates 7e-5, 9e-5 with NxMTransformer?}} \textcolor{red}{When you ran the experiments for them, we covered by 7e-5 and 9e-5 and up to rho 6e-3 and basically all the sweep directions worsened accuracy. Maybe increasing ADMM frequency would help but I haven't seen too much evidence of that...}
\fi
\end{table}

%% Double check GLUE score calculation method

We report the evaluation results for BERT in Table \ref{tab_glue-results} and make the following key observations. 

First, the pruning based method sparsifies weights of Transformer blocks but cannot explicit satisfy the underlying hardware constraints, e.g., the 4:2 sparsity. Although preserving the highest accuracy on downstream tasks (81.3 vs. 81.8 on average), the obtained sparse weights have a random structure of non-zero weights, which is inefficient to execute in modern hardware systems. As a result, the performance benefit with these unstructured sparsity based approaches is negligible, even when the pruning rate is high (e.g., 95\%)~\cite{structured-sparsity}.
% may even hurt the performance when executed as sparse matrix multiplication~\cite{any good citation here?}. 

Second, when it comes to NxM sparsity, \oursspace achieves an average score of 80.4, outperforming ASP by 1.6 points. In particular, we observe that for large tasks (MNLI, QNLI, SST-2), \oursspace performs comparably to ASP. However, \oursspace dramatically outperforms ASP for the small tasks (CoLA, STS-B, MRPC, RTE), increasing accuracy by 2.9 points on average. Since the smaller tasks can be more sensitive to random seed performance, we also performed a random seed sweep across 7 random seeds. \oursspace still outperformed ASP by 1.9 points across the tasks with fewer than 10,000 training examples when median was our metric of accuracy rather than maximum. This pattern suggests that while more data allows a less principled mechanism to recover accuracy, an explicit optimization approach that take the sparsity constraints into account would yield much 
better accuracy results when the downstream tasks have low data resources.  As a result, \oursspace retains 99\% of the accuracy of the unstructured sparsity (ADMM\textsubscript{Unstructured}) and 98.4\% of the uncompressed model (BERT\textsubscript{base}). Different from ADMM\textsubscript{Unstructured}, which suffers from expensive irregular memory accesses, our \oursspace method can effectively leverage the underlying Sparse Tensor Core and achieves inference speedups even with 50\% overall sparsity. 

% The performance of \oursspace can be broadly classified among two categories of tasks: the "small" tasks and the "large" tasks. On the large tasks (MNLI, QNLI, SST-2), \oursspace performs comparably to ASP. However, \oursspace dramatically outperforms ASP for the small tasks (CoLA, STS-B, MRPC, RTE), increasing accuracy by 2.9 points on average. This pattern suggests that while additional data can allow a less sophisticated mechanism to recover accuracy, when less data is available an explicit optimization approach better retains the accuracy of the original model.

% With \ours, the semi-structure sparsified BERT model can achieve comparable performance to the 

\subsection{Analysis Results}
\label{subsec:analysis-results}

In this section, we further investigate the performance gain of \oursspace and its impact to the downstream task accuracy with NxM sparsity. 

\iffalse
\minjia{This section is a bit hard to follow. @Connor, can you summarize the key messages for this section in a few sentences?}

\textcolor{red}{This section is basically just attempting to explain the metric we use for understanding ADMM, which we call similarity. It basically just is the proportion of weights that are in the semi-structured mask from ADMM iteration to the next. Most of the text is linking it to SAD, which SR-STE introduced as a principled metric, but their metric is basically just the absolute number of weights that change from iteration to iteration. The motivation for tying it to SAD is that I think it gives a little more credence to our similarity metric as something that is a valid mechanism for analysis. If that's unnecessary, I think we can just basically define our similarity metric and then discuss what we learn from it (which cuts most of the stuff).}
\fi

\begin{figure}[t]
    \centering
    \begin{subfigure}[t]{0.45\textwidth}
        \centering
        \includegraphics[width=\textwidth]{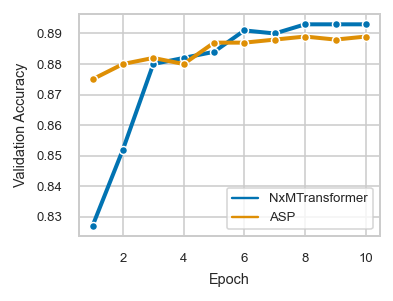}
        \caption{Validation accuracy of \oursspace and ASP networks on best configuration STS-B.}
        \label{fig_eval-accuracy}
    \end{subfigure}
    \hfill
    \begin{subfigure}[t]{0.45\textwidth}
        \centering
        \includegraphics[width=\textwidth]{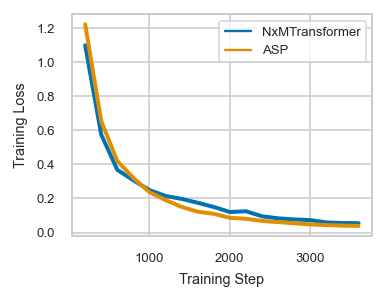}
        \caption{Training loss of the \oursspace and ASP networks on best configurations of STS-B.}
        \label{fig_training-loss}
    \end{subfigure}
    \caption{ASP and \oursspace on STS-B}
    \label{fig_admm-vs-asp}
\end{figure}

\paragraph{Validation accuracy improvement.} We first compare
the model trained with ASP and \ours. 
% Therefore, we create two configurations: \emph{NxMTransformer (Sparse)} and \emph{NxMTransformer (Dense)}.
% For \emph{NxMTransformer (Sparse)}, 
To evaluate NxMTransformer, we perform a hard prune of small weights at the end of every epoch, so we evaluate the model as if it has already sparsified. 
% For \emph{NxMTransformer (Dense)}, we perform the evaluation without the hard pruning step. 
As we can see in Figure~\ref{fig_admm-vs-asp}, \oursspace converges slower than ASP in the beginning of the fine-tuning. This is presumably because the initial model weights are heavily violating the hardware constraints, causing significant degradation when performing the pruning action. As the training moves forward, \oursspace is able to catch up and outperform ASP at around epoch 6. This is because by using ADMM, NxMTransformer trains the dense model to gradually converge to a sparse model that satisfies the provided constraints, so pruning weights would gradually have a smaller impact to model accuracy.
Since then, the validation accuracy of both ASP and \oursspace are still increasing, but ASP tends to plateau after 8 epochs, whereas \oursspace continues to increase, outperforming ASP by 0.5 point in the end. On the other hand, we also observe that ASP has slightly lower training loss towards the end (as shown in Figure~\ref{fig_training-loss}), indicating that ASP might be overfitting on the dev set (potentially due to the small amount of data). 

% \oursspace can require additional training time to achieve its best accuracy compared to ASP. Figure \ref{fig_admm-vs-asp} shows that will \oursspace reaches with .2 points if its maximium accuracy in 8 epochs, ASP requires just 5. However, ASP appears to overfit to the small training set more than \oursspace showing lower training losses despite consistently lower evaluation accuracy. While only demonstrated above using STS-B, this pattern was witnessed across the small tasks in general. We can further identify that the lack of a hard-pruning action helps maintain accuracy in the dense representation of the model, even as ADMM sparsifies it. 

\paragraph{Dynamism of sparse subnetwork masks.}

\begin{figure}[ht!]
    \centering
    \begin{subfigure}[b]{0.48\textwidth}
        \centering
        \includegraphics[width=\textwidth]{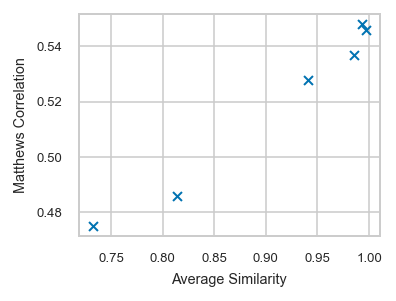}
        \caption{Similarity vs Matthews Correlation on CoLA (8.5k). Learning rate: 9e-5, Batch size: 32}
        % \minjia{Matthews Correlation is  the metric used by CoLA, right?}
        % \textcolro{red}{Yes, Matthews Correlation is just the CoLA accuracy metric.}
        \label{fig_cola-sad}
    \end{subfigure}
    \hfill
    \begin{subfigure}[b]{0.48\textwidth}
        \centering
        \includegraphics[width=\textwidth]{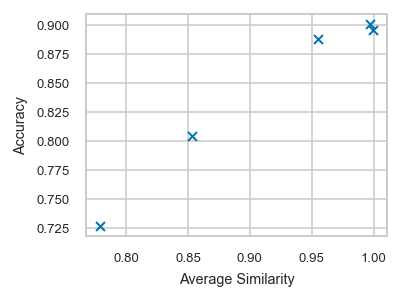}
        \caption{Similarity vs Accuracy on QNLI (108k samples). Learning: 3e-5, Batch size: 32}
        \label{fig_qnli-sad}
    \end{subfigure}
    \caption{Analysis of mask similarity and accuracy metrics for a subset of GLUE tasks. Different similarities are achieved through $\rho$ tuning.}
    \label{fig_sad-accuracy}
\end{figure}

The penalty parameter $\rho$ controls the balance between maximizing the accuracy of the model and reaching a sparse model. The larger that $\rho$ is, the greater the influence of the sparsity-inducing regularizer and the more quickly the model converges to a sparse solution. The trade-off represented by tuning $\rho$ manifests itself by values moving into and out of the sparse subnetwork mask between ADMM iterations. Since the sparsity is induced more slowly by a small $\rho$, a parameter is more likely to be included when the second sub-problem is solved. Empirically, we find that frequently moving values into and out of the sparse subnetwork mask results degrades the ultimate sparse networks accuracy. In Figure \ref{fig_sad-accuracy}, $\rho$ is tuned to achieve different values of similarity, which is calculated as the fraction of values that remain in the sparse subnetwork mask from one ADMM iteration to the next. For both CoLA and QNLI a clear correlation between average similarity and accuracy exists until approximately 99\% similarity, where the strength of the regularizer is over-weighted against the training loss and accuracy begins to degrade.

\begin{figure}[!ht]
    \centering
    \includegraphics[width=0.8\textwidth]{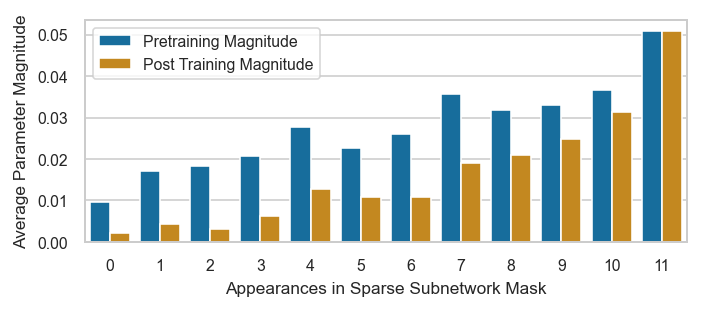}
    \caption{Comparison of average parameter value before and after fine-tuning on a 10-epoch STS-B experiment (learning rate: 5e-5, batch size: 16, $\rho$: 1e-3) based on the number of times it was present in the sparse subnetwork mask.}
    \label{fig_decay}
\end{figure}

Inspecting the values of weights that undergo swapping illustrates why higher dynamism in the sparse subnetwork mask incurs an accuracy penalty. Figure \ref{fig_decay} shows a parameter outside of the sparse subnetwork mask for just a single iteration will decrease in magnitude by approximately 15\% from its initial magnitude. A second iteration further increases this penalty to 25\%. This is in contrast to parameters that remain in the sparse mask for the entirety of the optimization process and retain all of their magnitude. The philosophy behind training from a pretrained model is to retain the information of that process. Large changes in parameter magnitude are destructive to that pretrained information because the parameter only partially reflects that learned information.

\subsection{When NxM Sparsity Meets Knowledge Distillation} \label{sec_distilbert-results}

Knowledge distillation has been proven to be another promising technique to compress a large model and also yield models with regular and dense structures. However, there have been few studies on the sparsity of knowledge distilled models, especially in the context of transfer learning and pre-trained language models. On first sight, it may seem that once distilling a large model into a smaller model, there will be less redundancy in the model, where sparsification might hurt model accuracy significantly. In this section, we investigate how \oursspace $ $affects KD compressed models. We apply \oursspace to a student model obtained through DistilBERT~\cite{distilbert}, which is a 6-layer BERT model with 768 hidden dimension size. The results are shown in Table~\ref{tab_distilbert}, and we make the following observations. 

\begin{table}[!ht]
\caption{
The dev set results on the GLUE benchmark with knowledge distillation. The results show \oursspace retains 97.6\% of the DistilBERT model.
}
\iffalse
\minjia{It would be better to add another column that shows the number of parameters (\#Params) of these models. The first two models have 66.6M parameters, and the third one 45.3M.}
\minjia{Also, this result shows the benefit of NxM sparsity but does not show the benefit of our ADMM-based approach. I still think it would be better to having either ADMM\textsubscript{unstructured} or APS results on DistilBERT.}
\fi
\label{tab_distilbert}
\resizebox{\textwidth}{!}{
\begin{tabular}{@{}lccccccccc@{}} \toprule
Model               & \# Params          & \multicolumn{7}{c}{Task} & Average \\ \cmidrule(r){3-9}
                    &                   & MNLI (m/mm)   & SST-2     & QNLI  & CoLA  & STS-B & MRPC  & RTE   &           \\ \midrule
\oursspace (4:2 BERT$_{12}$) & 66.6M          & {82.3/83.4}     & \textbf{92.3}      & \textbf{90.4}  & \textbf{55.3}  & \textbf{89.3}  & \textbf{90.8}  & \textbf{68.6}  & \textbf{80.5}      \\
DistilBERT (BERT$_{6}$)         & 66.6M                     & \textbf{82.4}/82.5     & 90.9      & 89.1  & 53.4  & 86.6  & 89.6  & 63.5  & 78.5      \\ \midrule
DistilBERT-\ours   & 45.3M    & \textbf{80.7/81.2}     & \textbf{90.5}      & \textbf{87.5}  & \textbf{50.1}  & \textbf{87.1}  & \textbf{88.7}  & \textbf{59.2}  & \textbf{76.6}      \\ \bottomrule
\end{tabular}}
% \minjia{TODO: Our results are better than DistilBERT. Need to talk about the comparison between NxMTransformer and DistilBERT, which compares semi-structured sparsity and structured compression. }
\end{table}

% The most simple form of knowledge distillation is that of DistilBert \cite{distilbert}. DistilBert performs logit-based distillation only during the pretraining process. Since the ADMM approach only integrates with the fine-tuning process, the ADMM methodology is identical for both DistilBert and BERT\textsubscript{BASE}.

% While the dimensionality of the model may be reduced, the relatively low amounts of sparsity exploited by semi-structured sparsity are still prevalent in fully connected layers.

First, despite DistilBERT and \oursspace have the same number of parameters, \oursspace  achieves 2 points higher accuracy on average than DistilBERT, which indicates that removing Transformer layers from the BERT model is more detrimental to the model accuracy and \ours's semi-structured approach captures redundancy (intra-layer) much more efficiently.  
% However, the expected performance gains from \oursspace are slightly smaller than DistilBERT. \oursspace retains more element-wise operations, and semi-structured scaling isn't nearly as efficient as layer pruning.

Second, \oursspace retains 97.6\% of the accuracy of the dense DistilBERT model. While slightly worse than the retained accuracy ratio for BERT\textsubscript{base} (98.4\%) itself, this indicates that while the depth of dimensionality of the model may be reduced, the relatively low amounts of sparsity exploited by semi-structured sparsity are still prevalent in fully connected layers. The result also seems to suggest the potential existence of a winning ticket even in highly compressed BERT model~\cite{lottery-bert}.

% it is still comparable despite the shallower depth and lower accuracy of DistilBERT. Similar to BERT\textsubscript{BASE}, \oursspace does a better job of retaining performance on the smaller tasks than the larger tasks.

More recently, knowledge distillation techniques such as TinyBert \cite{tinybert} and MiniLM \cite{structured-sparsity} leverage fine-grained knowledge transfer to help student better mimic teacher's behavior. As our method is largely orthogonal to how knowledge gets transferred between teacher and student, we expect the effectiveness on \oursspace as witnessed on DistilBERT should apply to models distilled through these more advanced techniques as well and will leave more extensive studies as future work.

% Since NxMTransformer only appends to the existing training loss, the presence of additional losses from attention maps, feature maps, or teacher logits meaningfully alters neither the behavior of ADMM nor the propagation of the losses from the teacher models. \oursspace also easily enables exemption of parameters from sparsification---learned projection layers can remain dense. Given \oursspace introduces minimal accuracy degradation as model depth decreases and the low barrier to integrating the additional distillation losses, the effectiveness on \oursspace as witnessed on DistilBERT should apply to these more advanced distillation techniques as well. 

\section{Conclusion} \label{sec_conclusion}

Semi-structured sparsity can improve runtime resource efficiency without large penalties in model performance. This work demonstrates the effectiveness of a low-overhead ADMM approach to introduce NxM semi-structured sparsity for large pretrained natural language models. Furthermore, \oursspace is an orthogonal optimization to existing compression techniques, such as knowledge distillation and reduced precision inference. However, NxMTransformer is limited in that it is not a lossless compression technique and does introduce an accuracy gap. Furthermore, it is untested on emerging pretrained Transformer representations for vision tasks and it is unclear how it would transfer to this emerging domain. 

\section{Negative Societal Effects} \label{sec_societal-effects}

NxMTransformer exposes two key avenues for negative societal impacts. First, since NxMTransformer is designed to inherit from a pretrained model representation, it inherits any societal-level biases that may be embedded in the parent model. Previous work has identified that BERT models do encode both gender bias \cite{measuring-in-word} and bias against people with disabilities \cite{biases-in-nlp-models}. NxMTransformer does not specifically attempt to combat these biases and downstream tasks fine-tuned with NxMTransformer will inherit them as well. The second potential source of negative societal impacts is due to the act of pruning itself. Hooker, et al. \cite{characterizing-bias} identify that for convolutional neural networks, pruning can disproportionately reduce accuracy of lower frequency output examples. Although the model design for CNNs is different from that of Transformers, it is unlikely this alone would mitigate this source of network bias. These sources of bias can introduce real-world harms as fine-tuned natural language models are increasingly used for online content moderating, brand sentiment, career matching, and other human-facing algorithms that can affect livelihoods. 

\begin{ack}
We thank the anonymous NeurIPS reviewers for their constructive comments. This work was supported by the National Science Foundation under grants CCF-1823005 and an NSF CAREER Award (CNS-1750760).
\end{ack}

\newpage

\bibliography{neurips_2021}
\bibliographystyle{plain}

%%%%%%%%%%%%%%%%%%%%%%%%%%%%%%%%%%%%%%%%%%%%%%%%%%%%%%%%%%%%
\section*{Checklist}

\begin{enumerate}

\item For all authors...
\begin{enumerate}
  \item Do the main claims made in the abstract and introduction accurately reflect the paper's contributions and scope?
    \answerYes{}
  \item Did you describe the limitations of your work?
    \answerYes{See Section \ref{sec_conclusion}}
  \item Did you discuss any potential negative societal impacts of your work?
    \answerYes{See Section \ref{sec_societal-effects}}
  \item Have you read the ethics review guidelines and ensured that your paper conforms to them?
    \answerYes
\end{enumerate}

\item If you are including theoretical results...
\begin{enumerate}
  \item Did you state the full set of assumptions of all theoretical results?
    \answerNA{No theoretical results.}
	\item Did you include complete proofs of all theoretical results?
    \answerNA{No theoretical results.}
\end{enumerate}

\item If you ran experiments...
\begin{enumerate}
  \item Did you include the code, data, and instructions needed to reproduce the main experimental results (either in the supplemental material or as a URL)?
    \answerYes{Training scripts and model code are included in the supplementary material.}
  \item Did you specify all the training details (e.g., data splits, hyperparameters, how they were chosen)?
    \answerYes{See Section \ref{sec_results}, Appendex \ref{sec_admm-hyperparameter-tuning}}
	\item Did you report error bars (e.g., with respect to the random seed after running experiments multiple times)?
    \answerNo{}
	\item Did you include the total amount of compute and the type of resources used (e.g., type of GPUs, internal cluster, or cloud provider)?
    \answerYes{See Section \ref{sec_results}}
\end{enumerate}

\item If you are using existing assets (e.g., code, data, models) or curating/releasing new assets...
\begin{enumerate}
  \item If your work uses existing assets, did you cite the creators?
    \answerYes{See Section \ref{sec_results}}
  \item Did you mention the license of the assets?
    \answerYes{See Section \ref{sec_results}}
  \item Did you include any new assets either in the supplemental material or as a URL?
    \answerNo{No new assets introduced in the work.}
  \item Did you discuss whether and how consent was obtained from people whose data you're using/curating?
    \answerNo{}
  \item Did you discuss whether the data you are using/curating contains personally identifiable information or offensive content?
    \answerNo{}
\end{enumerate}

\item If you used crowdsourcing or conducted research with human subjects...
\begin{enumerate}
  \item Did you include the full text of instructions given to participants and screenshots, if applicable?
    \answerNA{No human subjects.}
  \item Did you describe any potential participant risks, with links to Institutional Review Board (IRB) approvals, if applicable?
    \answerNA{No human subjects.}
  \item Did you include the estimated hourly wage paid to participants and the total amount spent on participant compensation?
    \answerNA{No human subjects.}
\end{enumerate}

\end{enumerate}

%%%%%%%%%%%%%%%%%%%%%%%%%%%%%%%%%%%%%%%%%%%%%%%%%%%%%%%%%%%%
\newpage
\appendix

\section{Appendix}

\subsection{Scheduling ADMM Iterations}
\label{sec_admm-hyperparameter-tuning}

ADMM can converge effectively with as few as 80 training steps in each ADMM iteration. For example, RTE (2.5k training samples), ADMM successfully converges with a minibatch of 32 and one ADMM iteration per epoch. However, increasing the number of training steps between iterations can reduce reliance on a high learning rate. Note that a high learning rate is necessary to allow the optimizer to relatively quickly push larger parameters towards 0 in a reasonable number of training steps, since the practical parameter delta in a single training step is proportional to the product of the learning rate and $\rho$. Furthermore, a small learning rate reduces the effectiveness of the regularizer and decreases model similarity.

Experimentally, ADMM will achieve its maximum accuracy once 10 ADMM iterations have occurred. However, further optimizing, does not appear to harm model accuracy. While further training is typically not desirable for small tasks --- training is frequently extended for these tasks to have a sufficiently large training period each ADMM iteration --- for large tasks tens of ADMM iterations may be performed such that the fine-tune can continue for sufficient time. For example, a fine-tune on QNLI for just 3 epochs may perform nearly 50 ADMM iterations.

\begin{table}[!ht]
\caption{NxMTransformer Training Hyperparameters. Smaller tasks utilize larger learning rates and penalty parameters ($\rho$) since ADMM iterations for these tasks are much shorter (See Appendix \ref{sec_admm-hyperparameter-tuning}).}
\label{tab_admm-hyperparameters}
\centering
\begin{tabular}{@{}lcccc} \toprule
Tasks & Learning Rates & $\rho$ & Batch Size & Epochs \\ \midrule
MNLI, QNLI, SST-2 & 1e-5, 3e-5, 5e-5 & 4e-4, 1e-3, 3e-3 & 16, 32 & 5 \\
CoLA, STS-B, MRPC, RTE & 5e-5, 7e-5, 9e-5, 1e-4 & 3e-3, 6e-3, 1e-2 & 16, 32 & 10 \\
\end{tabular}
\end{table}

\end{document}